\documentclass[conference]{IEEEtran}
\IEEEoverridecommandlockouts
\usepackage{cite}
\usepackage{amsmath,amssymb,amsfonts}
\usepackage{algorithmic}
\usepackage{algorithm}
\usepackage{graphicx}
\usepackage{subfigure}
\usepackage{textcomp}
\usepackage{verbatim}
\usepackage{xcolor}
\usepackage{subcaption}
\usepackage[colorlinks=true, linkcolor=blue, citecolor=blue, urlcolor=blue]{hyperref}

\def\BibTeX{{\rm B\kern-.05em{\sc i\kern-.025em b}\kern-.08em
    T\kern-.1667em\lower.7ex\hbox{E}\kern-.125emX}}
\begin{document}

\title{PFed-Signal: An  ADR
Prediction Model based on Federated Learning\\
% {\footnotesize \textsuperscript{*}Note: Sub-titles are not captured for https://ieeexplore.ieee.org  and
% should not be used}
\thanks{This study is supported by the National Natural Science Foundation
of China (Grant No. 62472250, 62372266, 62472251).}
}

\author{\IEEEauthorblockN{Tao Li}
\IEEEauthorblockA{\textit{School of
Computer Science} \\
\textit{Qufu Normal University}\\
Rizhao, China \\
 litao\_2019@qfnu.edu.cn}
\and
\IEEEauthorblockN{Peilin Li}
\IEEEauthorblockA{\textit{School of
Computer Science} \\
\textit{Qufu Normal University}\\
Rizhao, China \\
Lipeilin20000619@163.com}
\and
\IEEEauthorblockN{Kui Lu}
\IEEEauthorblockA{\textit{School of
Computer Science} \\
\textit{Qufu Normal University}\\
Rizhao, China \\
lukui0609@163.com}
\and
\IEEEauthorblockN{Yilei Wang}
\IEEEauthorblockA{\textit{School of
Computer Science} \\
\textit{Qufu Normal University}\\
Rizhao, China \\
wang\_yilei2019@qfnu.edu.cn}
\and 
\IEEEauthorblockN{Junliang Shang$^*$}
\IEEEauthorblockA{\textit{School of
Computer Science} \\
\textit{Qufu Normal University}\\
Rizhao, China \\
shangjunliang110@163.com}
\and 
\IEEEauthorblockN{Guangshun Li}
\IEEEauthorblockA{\textit{School of
Computer Science} \\
\textit{Qufu Normal University}\\
Rizhao, China \\
guangshunli@163.com}
\and
\IEEEauthorblockN{Huiyu Zhou}
\IEEEauthorblockA{\textit{School of Computing and Mathematical Sciences} \\
\textit{University of Leicester}\\
Leicester, UK\\
Hz143@leicester.ac.uk}
}

\maketitle

\begin{abstract}

 The adverse drug reactions (ADRs) predicted based on the biased records in FAERS (U.S. Food and Drug Administration Adverse Event Reporting System) may mislead diagnosis online. Generally, such problems are solved by optimizing reporting odds ratio (ROR) or proportional reporting ratio (PRR). However, these methods that rely on statistical methods cannot eliminate the biased data, leading to inaccurate signal prediction. In this paper, we propose PFed-signal, a federated learning-based signal prediction model of ADR, which utilizes the Euclidean distance to eliminate the biased data from FAERS, thereby improving the accuracy of ADR prediction. Specifically, we first propose Pfed-Split, a method to split the original dataset into a split dataset based on ADR. Then we propose ADR-signal, an ADR prediction model, including a biased data identification method based on federated learning and an ADR prediction model based on Transformer. The former identifies the biased data according to the Euclidean distance and generates a clean dataset by deleting the biased data. The latter is an ADR prediction model based on Transformer trained on the clean data set. The results show that the ROR and PRR on the clean dataset are better than those of the traditional methods. Furthermore, the accuracy rate, F1 score, recall rate and AUC of PFed-Signal are 0.887, 0.890, 0.913 and 0.957 respectively, which are higher than the baselines.
\end{abstract}

\section{Introduction}
\label{sec:introduction}
\IEEEPARstart{A}{dverse} drug reactions (ADRs) are one of the significant threats to clinical medication safety \cite{r1,r2}. According to the World Health Organization (WHO), the proportion of hospitalizations due to severe ADRs worldwide ranges from 3.7\% to 16.8\%, with approximately 5\% of hospitalized patients dying from ADRs. This severe situation highlights the crucial role of ADR predicting and early warning systems in clinical decision-making. To provide more clinical data for drug safety signal detection, the U.S. Food and Drug Administration (FDA) established the FAERS (FDA Adverse Event Reporting System) database \cite{r3,r4}, the world's largest spontaneous reporting system, to collect adverse event reports. The reports are submitted by patients, healthcare professionals, and pharmaceutical companies. 

However, there are some under-reportings in FAERS which  may bias the (clean) data and the biased data may lead to an increase in the reporting odds ratio (ROR) and the proportional reporting ratio (PRR).
%However, there are some significant security issues for FAERS: (1) The data sources are susceptible to poisoning attacks or backdoor attacks where clean data is contaminated. (2) Some data is under the risk of DDoS (Distributed Deny of Service) attacks and phishing attacks. These attacks may lead to network service interruptions, system crashes and data loss. (3) Underreporting is widespread, and some severe ADRs are not identified or reported and thus not reflected in the database. All these attacks and under-reportings may bias the (clean) data, which leads to an increase in the bias of traditional signal strength detection metrics, such as the reporting odds ratio (ROR) and the proportional reporting ratio (PRR).
Finally, the biased RoR and PRR mislead drug systhesis, online diagnosis decisions and patient medication \cite{r5,r6}. Recall that both ROR and PRR rely on statistical assumptions. While the data quality of FAERS (such as under-reporting, duplicate reporting, and incorrect labeling) can lead to an underestimation of the denominator (the number of non-target drugs or non-ADR reports), thereby artificially inflating ROR and PRR values and generating false positive signals. For example, if the severe ADR of a certain drug is not included in statistics due to under-reporting, its ROR value may be falsely magnified due to a reduced denominator, misleading clinical practice to believe that the drug is riskier. Although traditional optimization methods (such as Bayesian correction and threshold adjustment)  partially alleviate statistical fluctuations, they cannot solve the structural bias of the data source itself \cite{r8}. Therefore, it is urgent to optimize the accuracy of ADR detection from the perspectives of data cleaning and multi-source integration to avoid the biased data caused by security issues.

\subsection{Related works}
The ADR prediction includes traditional and machine learning-based methods. Currently, traditional ADR prediction models typically include PRR-based and ROR-based methods. ADR prediction based on PRR \cite{r9,r10,r11} involves calculating the proportion of adverse event reports in databases such as FAERS \cite{r12}, VAERS \cite{r13}, and MedWatch \cite{r14}. 
%A higher proportion is directly proportional to the strength of the ADR. ADR prediction based on ROR \cite{r15} involves extracting adverse event reports, constructing a fourfold table, and calculating the ROR and its 95\% confidence interval. 
%By combining a preset threshold (such as ROR $\geq$2 and the lower limit of the confidence interval $\geq$ 1, while the number of reports $\geq$ 3), the ROR is a commonly used method to determine the association between a drug and its adverse reactions. The ROR assesses the strength of the association between a drug and an adverse event by analyzing the report data \cite{r16}. 
The ROR method is based on simple statistical calculations and is easy to apply. Thus, it is suitable for rapid screening of large-scale data \cite{r17}. However, it is susceptible to reporting bias \cite{r18,r19}, to have low sensitivity against rare events \cite{r20,r21}, and cannot handle complex drug interactions \cite{r22,r23}. 
%Although \cite{r9,r10,r11} are relatively accurate for small sample data, they depend on the quality and completeness of the report data. 

%Generally, reporting bias (such as underreporting, overreporting, and selective reporting) may lead to inaccurate results. The research by Scosyrev et al. precisely addresses these shortcomings by introducing causal inference tools, especially causal graphs, to systematically identify and handle bias \cite{r24}. Causal graphs can help identify potential confounding factors and reduce the impact of bias through structured methods \cite{r25,r26}. However, causal inference relies on high-quality input data, yet data from spontaneous reporting systems (such as FAERS) may have reporting bias, incompleteness, and errors, which can weaken the effectiveness of causal inference \cite{r25,r27}. These non-machine learning methods have inaccuracy issues when predicting ADR. They cannot reduce or eliminate reporting bias in the original adverse reaction database and can only predict ADR based on biased reports, resulting in inaccurate predictions \cite{r28}.
\begin{figure*}[t]
    \centering
\includegraphics[scale=0.3]{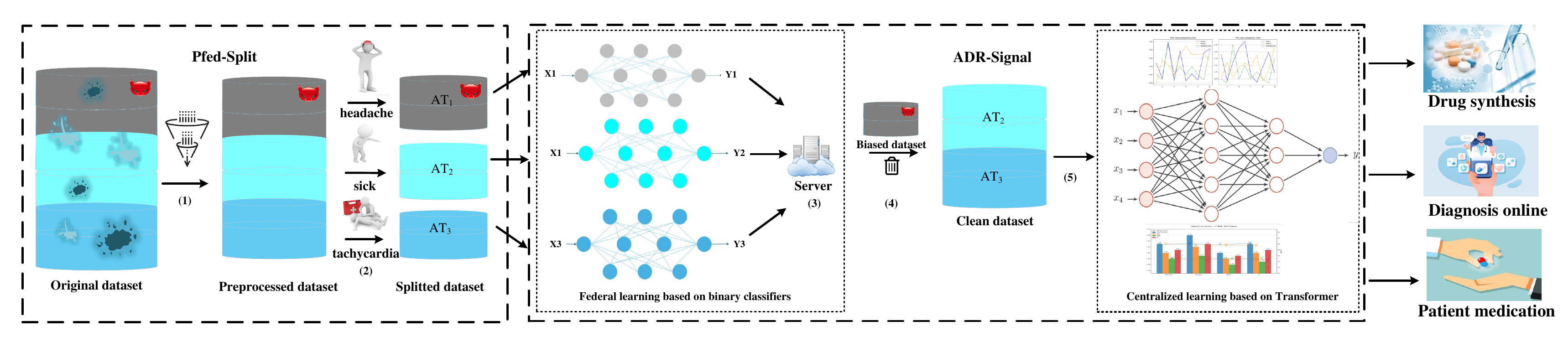} % 调整scale参数来改变图片大小，0.2表示缩小到原来的20%
    \caption{The framework of PFed-Signal: A comprehensive architecture for federated learning-based ADR prediction} % 为你的图添加说明
    \label{fig:motivations}
\end{figure*}

Machine learning-based ADR prediction commonly employs algorithms such as random forests (RF) \cite{r29}, support vector machines (SVM) \cite{r30}, and convolutional neural networks (CNNs) \cite{r31}. These methods identify the association signals between drugs and adverse reactions by mining the underlying patterns in medical data \cite{r32}. 
%By integrating the voting results of multiple decision trees and using feature importance assessment to screen key variables related to adverse reactions (such as patient symptoms, drug dosage, etc.), \cite{r29} enhances the stability and accuracy of predictions. By mapping low-dimensional nonlinear data to high-dimensional space through kernel functions and constructing the optimal classification hyperplane, \cite{r30} is suitable for distinguishing positive and negative signals of adverse reactions in small sample scenarios. Through local perception and weight sharing mechanisms, \cite{r31} automatically extracts deep patterns from medical data (such as sequential medication records or image features), and is particularly adept at handling high-dimensional sparse electronic health record data. 
In terms of advantages, random forests can effectively handle high-dimensional nonlinear data and have strong anti-overfitting capabilities \cite{r32}, SVM have excellent generalization performance in small samples and high-dimensional data \cite{r33}, and CNNs can capture complex data associations without the need for manual feature design \cite{r34}. However, these methods also have obvious limitations: the black-box nature of RF leads to poor interpretability of the results \cite{r35}, SVM are sensitive to the selection of kernel functions and parameter tuning and have high computational complexity \cite{r36}, and CNNs require massive amounts of labeled data and have significantly higher training costs than traditional methods \cite{r37}. Although the above methods have demonstrated strong technical advantages in the prediction of ADR, their performance is highly dependent on the quality of training data and cannot fundamentally eliminate the inherent systematic biases in the original medical database \cite{r38}, thus the predicted ADR inaccurate.

\subsection{Motivations and Contributions}

In summary, both traditional and machine learning-based methods suffer from the problem of biased data, which is particularly evident in spontaneous databases such as FAERS. The conclusions drawn on the biased data are at variance with the actual adverse reactions. Therefore, the biased data need to be eliminated before the ADR are predicted. Inspired by the mechanism of poisoning attack detection in Federated Learning (FL) \cite{r39}, biased data is similar to poisoning data. Recall that there is a large difference in the parameters between the model trained on normal data and poisoning data, which can be measured by the Euclidean distance between each local model and the global model. Therefore, federated learning can be used to find out and remove the biased data to get clean data. Consequently, the prediction of ADR can be made on the basis of the clean data, leading to a higher prediction accuracy of ADR.

In this paper, we propose PFed-Signal, which is a federated learning-based prediction model for adverse drug reactions (\textit{ref.} Fig.\ref{fig:motivations}), containing two components, Pfed-Split and ADR-Signal. The former divides FAERS into datasets adapted to federated learning, and the latter gets clean datasets through federated learning and trains ADR prediction models based on them. Specifically, Pfed-Split is a segmentation strategy that first preprocesses the original dataset data by cleaning, de-duplication, etc., to generate the preprocessed dataset (\textit{ref.} Step (1)). Then the preprocessed dataset is segmented into the split dataset (\textit{e.g.} $AT_1$, $AT_2$, $AT_3$) according to the adverse reactions (\textit{ref.} Step (2)), and the split dataset is used as the input of the ADR-Signal framework. The ADR-Signal framework trains the local model and global model on the basis of the split dataset (\textit{ref.} Step (3)). Then the Euclidean distance is used to measure the difference between the local model and the global model. When the Euclidean distance of a model exceeds a predefined threshold, we consider that the training dataset of the model contains biased dataset. Here we assume that $AT_1$ is the biased dataset and is deleted to generate a clean dataset (\textit{ref.} Step (4)). Based on the clean dataset, the ADR is predicted based on Transformers (\textit{ref.} Step (5)). The contributions of this paper are as follows.

(1) \textbf{PFed-Signal:} In this paper, we propose PFed-Signal, a federated learning-based ADR prediction model, which contains two novel components, Pfed-Split and ADR-Signal. 

(2)\textbf{ Pfed-Split:} It is an adverse reaction-based FAERS database segmentation strategy, which divides the original dataset into a split dataset (containing multiple ADR-based tables) based on adverse reactions through the steps of de-duplication, noise reduction, redundancy removal, and feature extraction. The split dataset supports the subsequent exclusion of biased data.

(3) \textbf{ADR-signal:} It is a Transformer-based adverse reaction prediction model. Specifically, ADR-signal first identifies and removes biased datasets in the split dataset using a personalized federated learning framework, and then merges the remaining ADR-based tables into the clean dataset. Finally, it predicts the ADR with the clean dataset based on Transformer.

(4)\textbf{ Evaluation.} Data are obtained from the FAERS database from Q1 2010 to Q3 2024, and the experimental environment uses the Python language combined with PyTorch in machine learning. The experimental results show that ROR and PRR based on the clean dataset are higher than those based on original dataset. In addition, compared with SVM, Bayesian Credible Propagation Neural Network method (BCPNN), and RF, PFed-Signal performs well in the prediction of ADR.

\section{The Framework of Pfed-Signal}
\subsection{Data source}
FAERS is a publicly accessible database maintained by the U.S. FDA. It documents adverse event reports related to drugs and medical devices, where the reports are submitted by patients, physicians, pharmaceutical companies, and other stakeholders. The FAERS data can be downloaded from the FDA's official website (\url{https://fis.fda.gov/extensions/fpd-qde-faers/fpd-qde-faers.html}) and is stored in structured tabular format. 

In this paper, a subset of data from the FAERS database, spanning from the first quarter of 2010 to the third quarter of 2024, is selected as the original dataset. It includes 10 types of adverse reactions, such as abnormal respiration, bone marrow suppression and aplastic anemia, and lower respiratory tract and pulmonary infections, along with 38 features including age, gender, weight, the year of patient reporting, patient outcomes, the presence of congenital malformations, and routes of drug administration. In total, there are 5,868 adverse event records in the original dataset. Recall that, as FAERS is a publicly available database, it may contain biased data. One of the primary tasks in this paper is to identify and address such biased data.

\subsection{The Workflow of Pfed-Signal}
In this paper, we propose a federated learning-based adverse reaction signal prediction model, named PFed-Signal, designed to identify the biased data and predict the intensity of adverse reaction signals, as illustrated in Fig. \ref{fig:motivations}. The core components of PFed-Signal are Pfed-Split and Pfed-Signal. The former divides the database into multiple subsets based on adverse reaction types, preparing for the identification of biased data (refer to Chapter \ref{pfedsplit}). The latter trains an adverse reaction prediction model on bias-free datasets (refer to Chapter \ref{adrsignal}). The basic workflow of PFed-Signal is as follows. Step (1): To clean data by removing redundant records, and perform feature selection on the \textit{original dataset} to generate a \textit{preprocessed dataset}. Step (2): To split the preprocessed dataset into \textit{split datasets} according to different ADRs, with each subset referred to as an \textit{ADR-based table}. For instance, the preprocessed dataset is divided into three ADR-based tables, namely $AT_1$, $AT_2$, and $AT_3$, based on the ADRs of headaches, sick, and tachycardia, respectively. It is important to note that biased data may not have been identified during the entire Pfed-Split process. Step (3): Utilizing a federated learning framework to train local binary classification models and a global model on $AT_1$, $AT_2$, and $AT_3$. After model training, the Euclidean distances between each local model and the global model are calculated. The local dataset corresponding to the local model with the largest Euclidean distance is defined as containing \textit{biased dataset} (\textit{e.g} $AT_1$). Step (4): To remove the biased dataset $AT_1$ from the original database to generate a \textit{clean dataset}, including $AT_2$, and $AT_3$. Step (5): To train an ADR prediction model on the clean dataset. It is noteworthy that PFed-Signal employs both federated learning and centralized learning training modes. Specifically, the framework trained using the split datasets is a federated learning framework based on binary classifiers, where each client possesses its own ADR-based table and trains a local model using a binary classifier locally. On the server side, in addition to aggregating training to generate a global model, the primary objective is to detect which ADR-based table contains biased data by calculating the Euclidean distances between local models and the global model, thereby generating the Clean dataset. On the other hand, the model trained using the Clean dataset is based on a centralized machine learning approach using Transformers, which is a multi-classifier. Its purpose is to comprehensively evaluate the intensity of adverse reaction signals by considering multiple types of adverse reactions. PFed-Signal not only eliminates biased data but also enhances the model's generalization ability, providing stronger support for drug safety monitoring and clinical decision-making. 

\section{The Component of Pfed-Split}
\label{pfedsplit}
\begin{figure}[t]
    \centering
\includegraphics[scale=0.23]{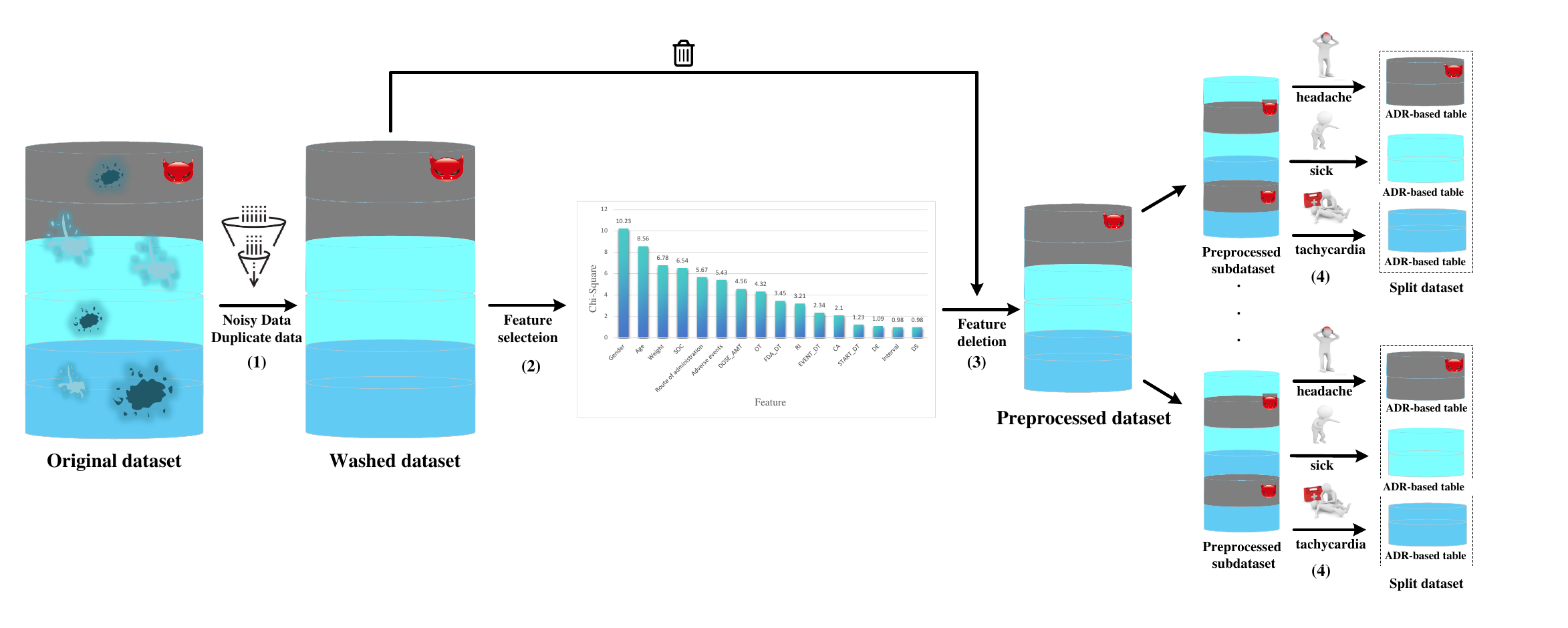} % 调整scale参数来改变图片大小，0.2表示缩小到原来的20%
    \caption{Workflow of Pfed-Split: Data preprocessing and segmentation for federated learning} % 为你的图添加说明
    \label{fig4}
\end{figure}

Recall that FAERS is not a distributed database and cannot be used directly for federated learning in ADR-Signal, which in turn identifies and deletes biased data. However, in PFed-Signal, if one wants to improve the accuracy of adverse reaction prediction, one must delete the biased dataset. Therefore, before identifying the biased dataset based on federated learning in ADR-Signal, it is necessary to generate a database that can participate in federated learning \cite{r19,r20}. This function is accomplished by PFed-Split, which consists of 4 steps as shown in Fig. \ref{fig4}.
%Contributions}

\subsubsection{Data Cleaning:} 
\begin{figure}[t]
    \centering
\includegraphics[scale=0.6]{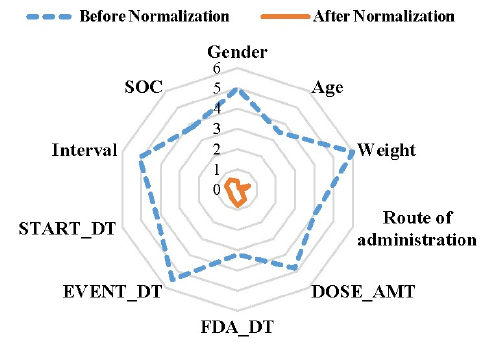} 
    \caption{Distribution of eigenvalues before and after data normalization: Enhancing data quality} 
    \label{fig5}
\end{figure}

This involves the removal of duplicate data and noisy data. After exporting the raw data from the FAERS, Python's pandas library is utilized. The $drop\_duplicates$ function is called, with key columns such as drug code, patient ID, and the date of adverse reaction occurrence specified to eliminate duplicate records and remove redundant information from the dataset. For noisy data, reasonable upper and lower limits are set for numerical data. By leveraging the \textit{pandas} library, abnormal values are screened, corrected, or deleted. Data normalization (as illustrated in Fig. \ref{fig5}) is then applied to compress the feature values into a new range, thereby eliminating the measurement bias in the original data. This process ensures that the numerical values of different features are more comparable, ultimately yielding a \textit{washed dataset}.

% \subsubsection{Feature Selection} {Following \cite{r41}, the Chi-square test is employed to screen features associated with adverse reactions. In this paper, the \textit{Chi2} function in Python is utilized to calculate the Chi-square value for each feature. Subsequently, significant features are selected based on these Chi-square values (with a threshold of Chi-square value greater than 3). Fig. \ref{fig6} presents the Chi-square values of some features in the Washed dataset, including Gender, Age, Weight, System Organ Class (SOC, indicating medication interval or relevant drug category grouping in the context of adverse reaction reporting), Route of administration, Adverse events, Preferred Term (DOSE\_AMT, representing the primary description of adverse events), Other Serious Outcomes (OT), FDA receipt date (FDA\_DT), Requirement for Intervention (RI), the date of the patient's first occurrence of an adverse event (EVENT\_DT), Whether congenital malformation (CA), Start date (START\_DT), Death indicator (DE), Duration interval of adverse reactions (Interval), and whether intervention is required (DS).}

% \begin{figure}[t]
%     \centering
% \includegraphics[scale=0.8]{fig6.pdf} 
%     \caption{Chi-square values of features: Selecting relevant features for ADR prediction} 
%     \label{fig6}
% \end{figure}

 \subsubsection{Feature Deletion}  This step removes redundant features from the Washed dataset, retaining only the significant features identified through feature selection. However, due to data missing during the upload process, some of these significant features contain \textit{Null }values. Therefore, records with \textit{Null} values in significant features are further deleted, ultimately yielding the \textit{preprocessed dataset}.
\subsubsection{Dataset Splitting by Type} It is important to note that the preprocessed dataset remains a centralized database, which does not meet the requirements of federated learning. Directly splitting it into different subsets based on adverse reaction (ADR) types would result in uneven data distribution across ADRs and would not align with the needs of the federated learning framework. To address this, the preprocessed dataset is first uniformly divided into several split subdatasets $Split_i$ ($i\in\{1,2,...n\}$), where $n$ is the number of the clients. All split subdatasets are included in a Split dataset $Split=\{Split_i\}$. Note that, each $Split_i$ encompass a variety of ADR data. Then each split subdataset $Split_i$ is further split into ADR-based tables  $Split_i=\{AT^i_j|j\in\{1,2,...m\}\}$, where $m$ is the number of ADRs.

\section{The Component of ADR-Signal}
\label{adrsignal}
Recall that, ADR-Signal consists of two phases: (1) Clean dataset based on federated learning, which uses the binary classifiers in federated learning to identify the biased data, which in turn generates the clean dataset, and (2) ADR prediction based on the Transformer, which trains the adverse reaction prediction model on the clean dataset.
\label{sec:guidelines}

\subsection{Clean dataset based on federated learning}
\begin{figure}[t]
\centering
\includegraphics[scale=0.1]{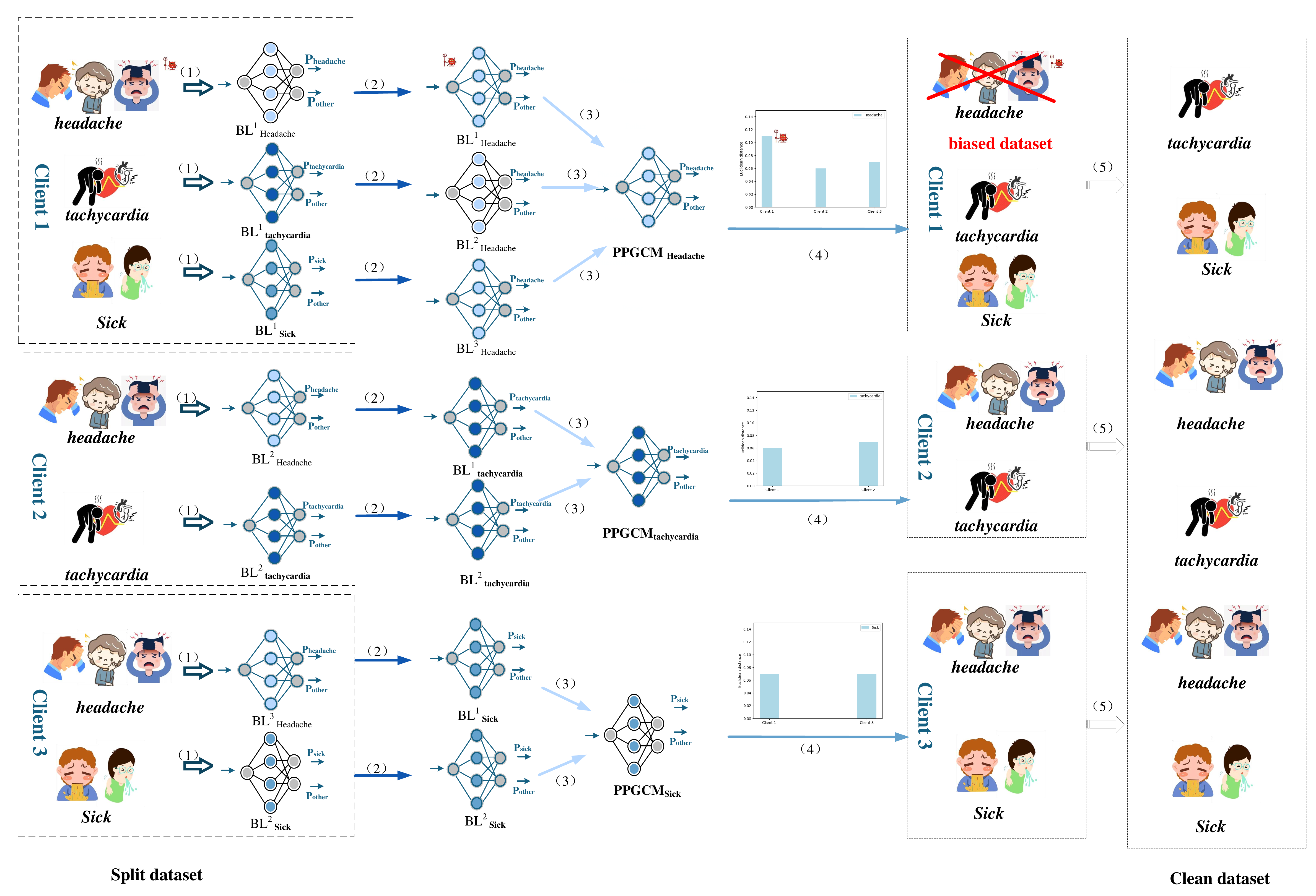} % 调整scale参数来改变图片大小，0.2表示缩小到原来的20%
    \caption{ Clean dataset based on federated learning: Identifying and removing biased data} % 为你的图添加说明
    \label{fig7}
\end{figure}
Inspired by the poisoning data detection methods based on federated learning \cite{posion}, this paper leverages federated learning to identify biased data and generate a clean dataset. In general, $client_i$ possesses a $split_i$, and trains corresponding local binary classifiers $LB^i_j$ based on $AT^i_j$ for each ADR $j$. Here, \( BL^{i}_{j} \) denotes a local binary classifier, where \( i \) represents client \( i \), and \( j \) represents the ADR.  At the server side, $LB^i_j$ are aggregated to $PPGCM_j$ according to the $j^{th}$ ADR. Subsequently, the Euclidean distances caculated between $LB^i_j$ and $PPGCM_j$. The $AT^i_j$ corresponding to $LB^i_j$, which is greater than the threshold, is deemed as biased data. After removing the biased data from the split subdatasets and aggregating the remaining split subdatasets, a Clean dataset is generated. The generation of the clean dataset involves 5 steps. Fig. \ref{fig7} presents an instance with 3 clients and 3 ADRs of headache, tachycardia and sick.
 Step (1): Each client trains binary classifiers locally based on its own ADR-based tables. For instance, $client_1$ has three ADR-based tables for headache ($AT^1_{headache}$), tachycardia ($AT^1_{tachycardia}$), and sick ($AT^1_{sick}$). $client_1$ trains three local binary classifiers:  $BL^1_{headache}$, $BL^1_{tachycardia}$ and $BL^1_{sick}$. Step (2): Each client uploads its local binary classifiers to the server for pre-aggregation. Step (3): The server performs pre-aggregation of the uploaded binary classifiers by category, generating a Pre-aggregated Partial Global Classifier Model (PPGCM) for each type of ADR. For example, considering three types of ADRs---headache, tachycardia, and sick---the server generates three corresponding PPGCMs: $PPGCM_{headache}$, $PPGCM_{tachycardia}$, and $PPGCM_{sick}$. Step (4): For each type of ADR, the Euclidean distances between each $BL^{i}_{j}$ and $PPGCM_j$ are calculated. If a local model's Euclidean distance is significantly higher than the threshold $\epsilon=4$, the training dataset $AT^i_j$ of $BL^i_j$ is considered as biased data. For instance, the Euclidean distance of $BL^1_{headache}$ is higher than $\epsilon$, then the ADR table  $AT^i_j$ is considered as a biased data. Step (5): The biased data $AT^i_j$ is removed from $split_1$, which is then merged with the $split_2$ and $split_3$ to form the Clean dataset.

% \begin{table}[t]
% \caption{The Euclidean Distances between Local and the Pre-aggregated Global Models}
% \centering
% \label{table}
% %\setlength{\tabcolsep}{3pt}
% \begin{tabular}{|l|c|c|c|}
% \hline
% & $PPGCM_{h}$
% & 
% $PPGCM_{t}$ & $PPGCM_{s}$\\
% \hline
% $BL^{1}_{h}$& 
% \textbf{4.7421}&/&/\\
% $BL^{2}_{h}$& 
% 2.9135& 
% / &/\\
% $BL^{3}_{h}$& 
% 2.9135& 
% /&/ \\
% $BL^{1}_{t}$&/&2.2452&/\\
% $BL^{2}_{t}$& 
% /& 2.3667 &/\\
% $BL^{1}_{s}$& 
% /& 
% /&2.1829\\
% $BL^{3}_{s}$& 
% /& 
% /&2.0688 \\
% \hline
% \multicolumn{3}{p{20pt}}{ }\\
% \end{tabular}
% \label{tab1}
% \end{table}

%  Table \ref{tab1} shows the Euclidean distance between the local binary classifiers and the pre-aggregated global model, which is used to measure the difference between each local model and the pre-aggregated global model. The bold one identifies the Euclidean distance of the local binary classifier is higher than the threshold $\epsilon=4$. For simplicity, in Table \ref{tab1}, ``$h$" denotes headache, ``$t$" denotes tachycardia and ``$s$" denotes sick. It's clear that the Euclidean distance of $BL^1_h$ and $PPGCM_h$ is higher than $\epsilon=4$, while others are not. Therefore, the training set of $BL^1_h$ is considered as biased data.

\subsection{ ADR prediction based on the Transformer}
\begin{figure*}[t]
    \centering
\includegraphics[scale=0.3]{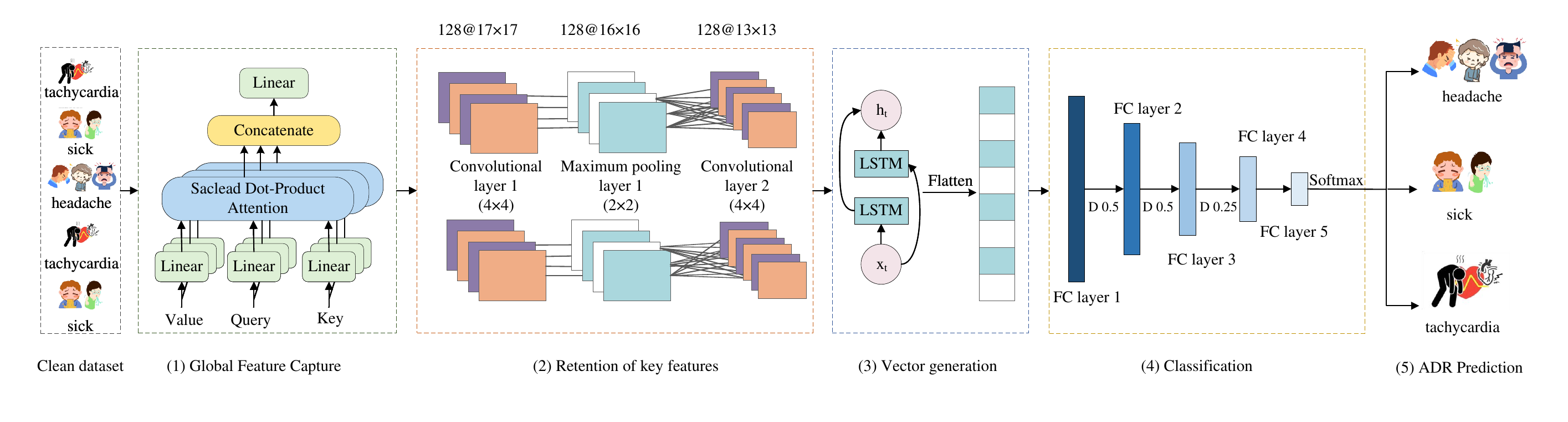} % 调整scale参数来改变图片大小，0.2表示缩小到原来的20%
    \caption{ADR prediction based on Transformer: Capturing complex signal correlations} % 为你的图添加说明
    \label{fig8}
\end{figure*}
ADR-Signal is a Transformer-based detection model. The self-attention mechanism in Transformer can effectively handle the complex correlations of signals and capture the global features of the adverse response signal, thus improving the detection accuracy of the adverse signal (as shown in Fig. \ref{fig8}).
(1) Global Feature Capture: Leverage multiple distinct attention heads within the Transformer architecture to capture diverse features from the clean dataset, thereby enhancing the model's expressiveness. The clean dataset undergoes three separate linear transformation layers to generate three vectors: Query, Key, and Value. These vectors are then divided into several ``heads,'' with each head having its own independent Query, Key, and Value matrices. The Scaled Dot-Product Attention operation is performed on each head, and the outputs from all heads are concatenated together. Finally, a linear layer is applied to fuse these outputs, yielding the final attention output vector.

(2) Retention of Key Features: Initially, a one-dimensional convolutional layer (Convolutional Layer 1 with a \(4 \times 4\) kernel) is employed to extract local features. Dimensionality reduction is achieved through a max-pooling layer. Subsequently, another one-dimensional convolutional layer (Convolutional Layer 2 with a \(4 \times 4\) kernel) is utilized to extract deep-level features, which are retained as the key features.

(3) One-Dimensional Vector Generation: The key features are subjected to sequential learning using an LSTM (Long Short-Term Memory) network. A forward LSTM processes the feature sequence from left to right, while a backward LSTM processes it from right to left, ultimately generating dynamic memory units \(h_t\) that enable the model to capture contextual information. The feature data is then flattened into a one-dimensional vector via a Flatten layer, facilitating the exploration of more complex sequential patterns for input into a fully connected layer.

(4) Signal Intensity Prediction: The one-dimensional vector is fed into a fully connected neural network (FCNN) for processing. It passes through five fully connected layers (FC Layer 1 to FC Layer 5) and three Dropout functions (\(D(0.5)\), \(D(0.5)\), \(D(2.5)\)) to progressively reduce the feature dimensionality. Finally, a softmax activation function is applied to classify the output, converting the raw output values into a probability distribution. The predicted category by the model is the one with the highest probability, enabling the classification and prediction of signal intensities for various adverse reactions.

Algorithm \ref{alg2} is the pseudocode of the PFed-Signal model: Inputs include the Split dataset $Split$, binary classifier configurations ($BC\_Config$) and neural network configuration ($NN\_Config$). The output is the  signal detection results ($signal\_predictions$). In Algorithm \ref{alg2}, the 
$clean\_dataset$, $biased\_dataset$ are first initialized (lines 4-6). Initialize a binary classifier $BC$ and a neural network model ($NN$) with $BC\_Config$ and $NN\_Config$ respectively (lines 7-8). Then $Client_i$ trains its local binary classifiers $LB^i_j$ on $AT^i_j$ and upload them to the server (lines 9-11). Note that it not necessarily for $Client_i$ to train $m$ binary classifiers. In fact, the number of $LB^i_j$ depends on the number of ADR types in $AT^i_j$. The sever collects the local binary classifiers and pre-aggregate them to find the biased data. Specifically, the server computes $PPGCM_j$ by calculating the average of $LB^i_j$ (line 13). Then the server evaluates the Euclidean distance between each $LB^i_j$ and $PPGCM_j$ (line 14). If the distance is lower than $\epsilon$, then the corresponding dataset $AT^i_j$ is added to the $clean\_dataset$. Otherwise, $AT^i_j$ is added to the $biased\_dataset$ (lines 15-17). In the following, we can predict the ADR with the clean dataset with $NN(clean\_dataset)$ and get a probability of the ADR with a sigmoid function (lines 19-20). Finally, the signals with $p\geq 0.9$ is appended to the prediction results, indicting there is a high ADR.  

\begin{algorithm}[t]

\caption{ADR-Signal}\label{alg2}
\begin{algorithmic}[1]
\STATE Input: $Split$, $\epsilon$, $BC\_Config$ $NN\_Config$ 
\STATE Output: Signal\_Predictions
\STATE \textbf{Initialize:}
\STATE \hspace{5pt} \textit{clean\_dataset} = []
\STATE \hspace{5pt} \textit{biased\_dataset} = []
\STATE \hspace{5pt} \textit{signal\_predictions} = []
\STATE \hspace{5pt} $BC= Build\_BinaryClassifier(BC\_Config)$
\STATE \hspace{5pt} $NN= Build\_NeuralNetwork(NN\_Config)$
\STATE \textbf{For each $Client_i$}
\STATE  \hspace{5pt}  $LB^i_j=BC(AT^i_j)$
\STATE   \hspace{5pt} Upload $LB^i_j$ to the server
\STATE \textbf{For the server}
\STATE \hspace{5pt}   $PPGCM_j=Average(LB^i_j)$
\STATE \hspace{5pt}   $Distance^i_j(LB^i_j,PPGCM_j)$
\IF{$Distance^i_j<\epsilon$}
    \STATE $clean\_dataset \leftarrow AT^i_j$
\ELSE       
\STATE $biased\_dataset \leftarrow AT^i_j$
\ENDIF
\STATE \textbf{The prediction of ADR}
\STATE \hspace{5pt}  $output = NN(clean\_dataset)$
\STATE \hspace{5pt}  $p= \mathrm{Sigmoid}(output)$

\IF{$p \geq 0.9$}
    \STATE $signal\_predictions.\mathrm{append}(p)$

\ENDIF
\STATE \textbf{return $signal\_predictions$}
\end{algorithmic}
\label{alg2}
\end{algorithm}

\section{Evaluation}
\subsection{Experiment settings}
(1) Environment: The experiments are conducted using Python in conjunction with PyTorch on a machine learning platform. The computational resources included a 2.50 GHz 8255C CPU, 16GB of RAM, and an NVIDIA RTX 3080 GPU card.

(2) Dataset: The dataset spans from Q1 of 2010 to Q3 2024 of FAERS database, encompassing 10 types of adverse reactions and 38 features such as age, gender, weight, and drug administration routes. The original dataset comprises 5,868 adverse event records.

(3) Performance Metrics: Several key performance metrics are employed to evaluate the effectiveness of the models, including the F1 score, accuracy, recall, and AUC (Area Under the Curve). These metrics are crucial for assessing the models' performance to predict ADR singals.

(4) Baseline Models: We compare the performance of PFed-Signal with three baseline models: Support Vector Machine (SVM) \cite{r43}, Bayesian Credible Propagation Neural Network (BCPNN) \cite{r44} and Random Forest (RF) \cite{r45}.

% Support Vector Machine (SVM): A supervised learning algorithm \cite{r43} primarily used for classification and regression analysis, renowned for its good generalization ability on small sample sizes and high-dimensional data.

% Bayesian Credible Propagation Neural Network (BCPNN): A neural network model based on Bayesian theory \cite{r44} that combines the learning capabilities of neural networks with the uncertainty quantification of Bayesian statistics, making it particularly suitable for handling data with uncertainty.

% Random Forest (RF): An ensemble learning method \cite{r45} that improves model accuracy and stability by constructing multiple decision trees and combining their predictions, suitable for handling high-dimensional features and large datasets.

%While PFed-Signal shares similarities with these baseline models in terms of processes such as feature extraction and model training, its advantages lie in its adoption of a federated learning framework for data processing. Additionally, PFed-Signal identifies and eliminates biased data by leveraging Euclidean distance, thereby generating a clean dataset that contributes to enhancing the accuracy of adverse drug reaction signal prediction. By comparing PFed-Signal with these three baseline models, we can demonstrate its superior performance in predicting adverse drug reactions.
\subsection{ Performance of PFed-Signal}
\subsubsection{The metrics of PFed-Signal}
\begin{figure*}[t]
    \centering
\includegraphics[scale=0.15]{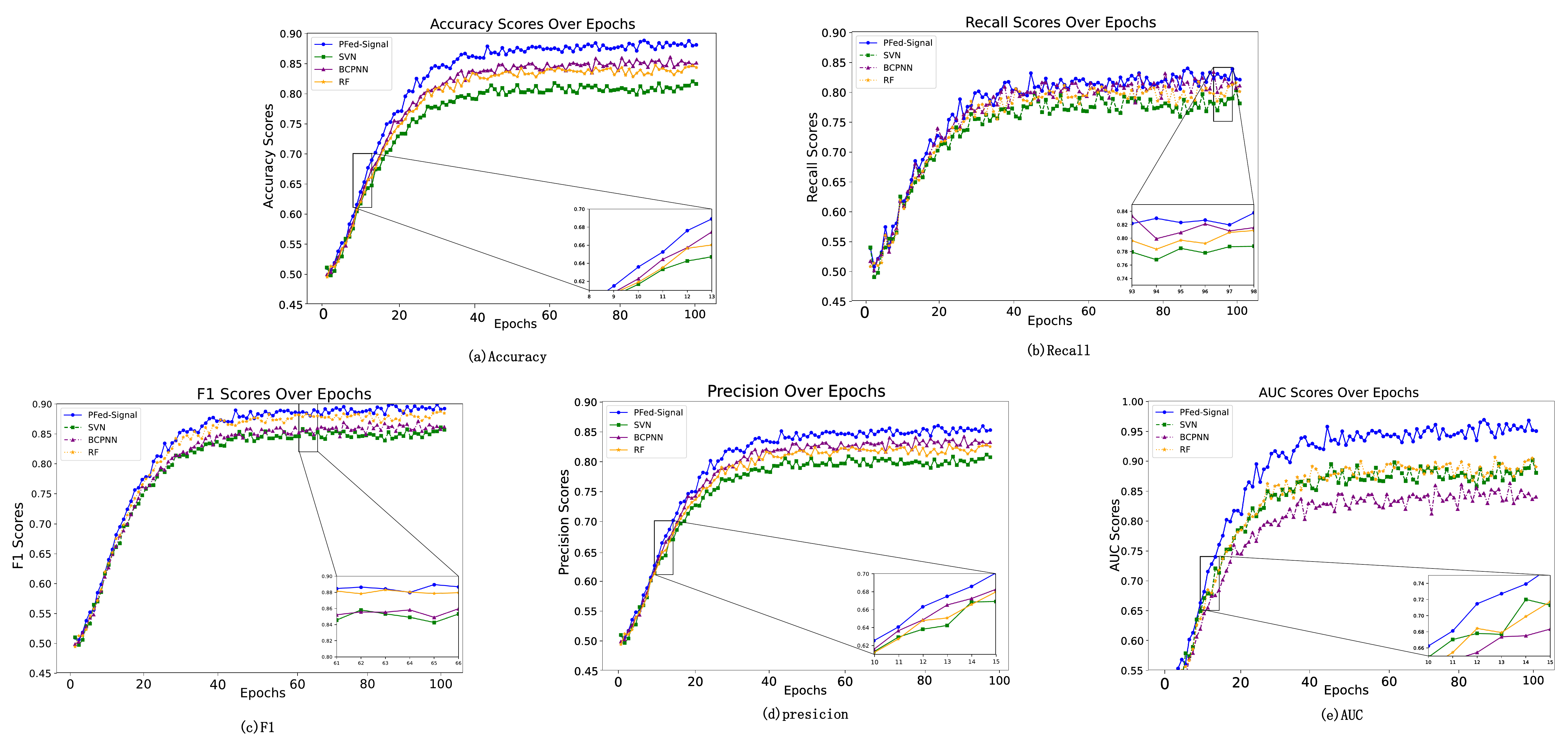} % 调整scale参数来改变图片大小，0.2表示缩小到原来的20%
    \caption{Performance metrics of PFed-Signal and baselines: Demonstrating superior prediction accuracy} % 为你的图添加说明
    \label{fig9}
\end{figure*}

The metrics of accuracy, AUC, precision, F1 score, and recall of PFed-Signal are given in Fig. \ref{fig9}, where the training dataset is clean dataset and the number of training epochs is 100. Fig. \ref{fig9} (a)(b)(c) shows that from the $10^{th}$ round of training, the accuracy, AUC, and precision of PFed-Signal gradually exceed the baselines (SVM, BCPNN and RF). When the number of training epochs reaches 100, the accuracy, AUC, and precision of PFed-Signal reaches the maximum of 0.88, 0.95, and 0.85, respectively, which are significantly higher than those of baselines. Fig. \ref{fig9} (d)(e) shows that at the beginning of the training epochs PFed-Signal's F1 score and recall are better than other models, but their advantages are not stable. When the training epochs are 60 and 90 respectively, the F1 score and recall are stable and higher than other models. When the training epochs are 100, the F1 score and recall of PFed-Signal reaches the maximum of 0.89 and 0.82 respectively, which is significantly higher than those of baselines. In Fig. \ref{fig9}, PFed-Signal significantly outperforms the baselines in all key metrics.

 \begin{table}[t]
\caption{The Matrix of PFed-Signal and Baselines}
\centering
\scriptsize
\label{table}
\begin{tabular}{|c|c|c|c|c|}
\hline
Metric& SVM
& 
BCPNN & RF&PFed-Signal\\
\hline
Accuary& 
0.813($\downarrow$7.4\%)&0.855($\downarrow$3.2\%)&0.841($\downarrow$4.6\%)&\textbf{0.887}\\
AUC&0.884($\downarrow$7.0\%)&0.845($\downarrow$10.9\%)&0.891($\downarrow$6.3\%)&\textbf{0.954}\\
Precision& 
0.807($\downarrow$4.4\%)& 
0.834($\downarrow$1.7\%)&0.822($\downarrow$2.9\%)&\textbf{0.851}\\
F1 scores& 0.854($\downarrow$4.1\%)
& 0.862($\downarrow$3.3\%) & 0.885($\downarrow$1.0\%)
 &\textbf{0.895}\\
Recall& 
0.784($\downarrow$3.6\%)& 
0.818($\downarrow$0.02\%)&0.801($\downarrow$1.9\%)& \textbf{0.820}\\
\hline
\multicolumn{3}{p{15pt}}{ }\\
\end{tabular}
\label{tab2}
\end{table}

Table \ref{tab2} presents the average values of accuracy, AUC, precision, F1-score, and recall for PFed-Signal and the baseline models (SVM, BCPNN, RF) after 100 epochs of training. It is evident that the PFed-Signal model significantly outperforms the baselines. PFed-Signal achieves an accuracy of 0.887, while SVM, BCPNN, and RF achieve accuracies of 0.813, 0.855, and 0.841, respectively. The AUC value of PFed-Signal is 0.954, which is 7\% higher than that of SVM (0.884). The precision of PFed-Signal is 0.851, compared to 0.807, 0.834, and 0.822 for SVM, BCPNN, and RF, respectively. The F1-score of PFed-Signal is 0.895, whereas those of SVM, BCPNN, and RF are 0.845, 0.862, and 0.885, respectively. The recall value of PFed-Signal is 0.820, which is notably higher than those of the baselines. As shown in Table \ref{tab2}, PFed-Signal demonstrates superior performance in all key metrics compared to the other three models.

\begin{figure}[t]
    \centering
\includegraphics[scale=0.15]{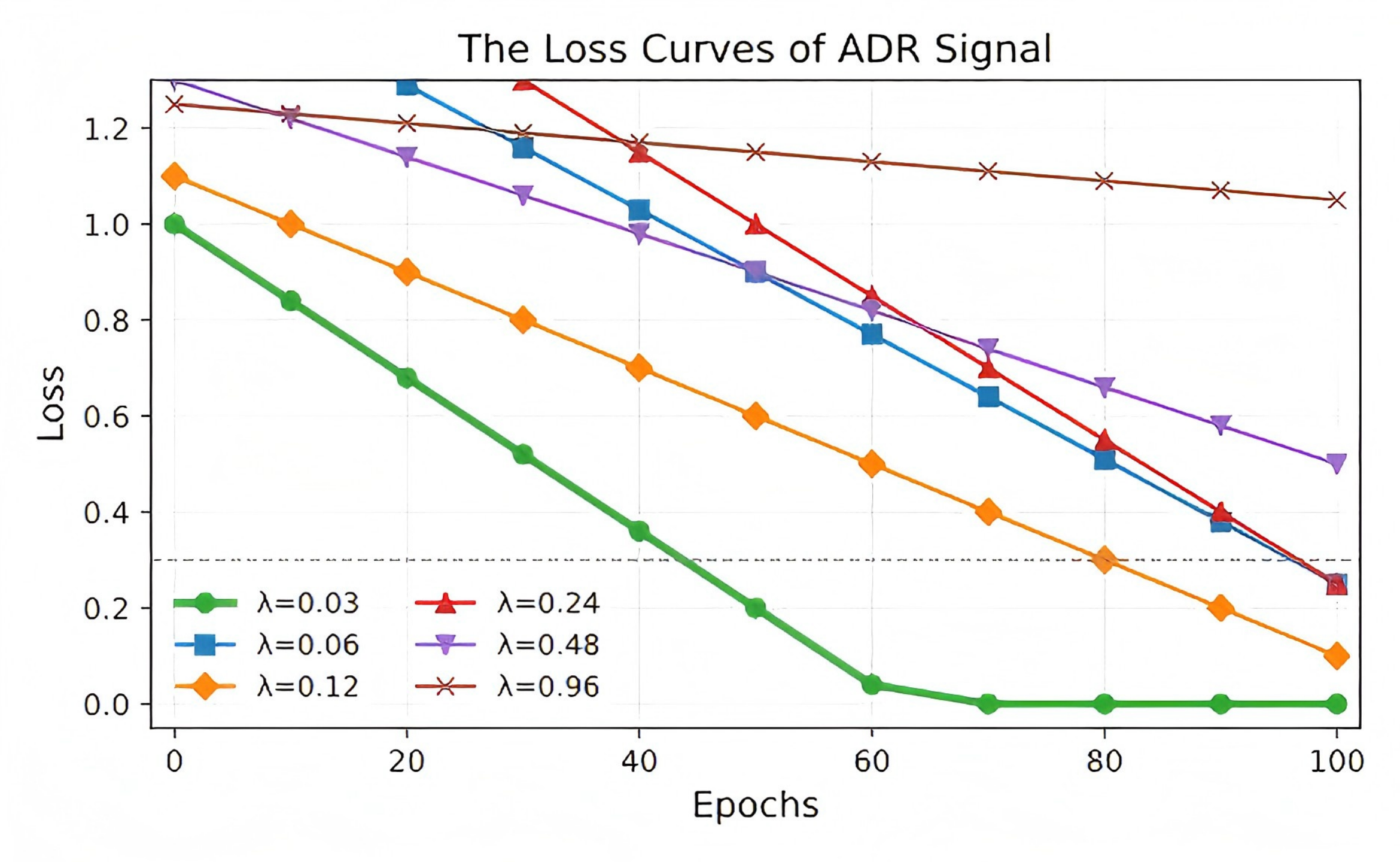} % 调整scale参数来改变图片大小，0.2表示缩小到原来的20%
    \caption{ Loss curves of ADR-Signal: Evaluating model optimization} % 为你的图添加说明
    \label{figure10.pdf}
\end{figure}

To verify whether the PFed-Signal model is overfitting, we present the loss values of PFed-Signal at each epoch. In fact, these are the loss values of the ADR-Signal model, which is a Transformer-based model for predicting the intensity of adverse drug reaction signals, as illustrated in Fig. \ref{figure10.pdf}. The horizontal axis represents the epochs, and the vertical axis represents the model's loss value, with $\lambda$ denoting the learning rate. Following the work \cite{r47}, 
we selected $\lambda_{1} = 0.03$, $\lambda_{2} = 0.06$, $\lambda_{3} = 0.12$, $\lambda_{4} = 0.24$, $\lambda_{5} = 0.48$, and $\lambda_{6} = 0.96$ (a geometric progression with common ratio 2).  As shown in Fig. \ref{figure10.pdf}, the loss decreases monotonically with training iterations under all tested rates, indicating continual model optimization. Crucially, we observe a clear performance degradation when the learning rate is doubled ($\lambda_{1} = 0.03 \rightarrow \lambda_{2} = 0.06$), allowing us to pinpoint the optimal value: it avoids both the sluggish convergence of overly small rates and the excessive update magnitudes caused by overly large ones. This confirms that an appropriately chosen learning rate enables more effective loss reduction, whereas values that are either too small or too large can precipitate overfitting. During the initial hyper-parameter search, we selected $0.03$ rather than nearby values such as $0.029$ or $0.031$ to preserve the regularity of the geometric progression and to prioritize rapid identification of the effective range over minor fluctuations, thereby balancing experimental efficiency with interpretability. Consequently, $\lambda_{1} = 0.03$ is employed throughout our model training.
\subsubsection{Identification Rate of Biased Data}
This paper employs the confusion matrix as a metric to evaluate the accuracy of ADR-Signal in identifying biased data. Recall that a confusion matrix presents the correspondence between actual and predicted classes, along with the number of samples in each class, in a tabular form \cite{r46}. 
\begin{comment}In Fig. \ref{fig11}, ``Biased'' represents the biased data in the database, which is typically caused by false positives or misreporting, while ``Unbiased''' denotes data without such biases. The primary objective of PFed-Signal is to identify and remove biased data to ensure the reliability and accuracy of ADR prediction model.\end{comment}

In the Original dataset, there are 44 genuine biased data points labeled as ``Biased'' and 5,824 non-biased data points labeled as ``Unbiased''. After identification using Euclidean distance, 43 data points were marked as ``Biased'', and 5,825 data points were marked as ``Unbiased''. Consequently, 43 samples of ``Biased'' data were correctly identified, yielding an impressive accuracy rate of 0.97 in identifying biased data.

\begin{comment}
\begin{figure}[t]
    \centering
\includegraphics[scale=0.58]{fig11.pdf} % 调整scale参数来改变图片大小，0.2表示缩小到原来的20%
    \caption{Confusion matrix of biased dataset identification: Assessing the accuracy of bias detection} % 为你的图添加说明
    \label{fig11}
\end{figure}
\end{comment}

\subsubsection{The optimization of ROR and PRR}
\begin{figure}[t]
    \centering
\includegraphics[scale=0.55]{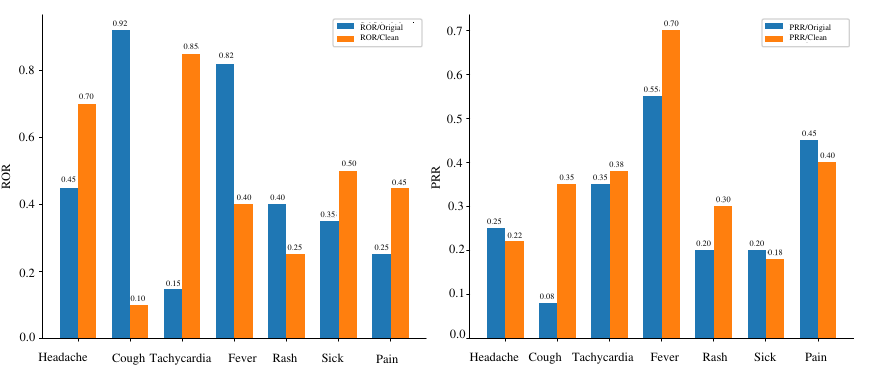} % 调整scale参数来改变图片大小，0.2表示缩小到原来的20%
    \caption{Comparison of ROR and PRR using original and clean datasets: Highlighting the impact of data cleaning} % 为你的图添加说明
    \label{fig12}
\end{figure}

Recall that after the identification of biased data, a clean dataset is generated. Based on this clean dataset, the ADR prediction model exhibits better performance. Moreover, the clean dataset can also be applied to the calculation of traditional ROR and PRR indicators, demonstrating the good compatibility of PFed-Signal. Fig. \ref{fig12} presents the values of ROR and PRR calculated on the Original dataset and the clean dataset. PRR/original and ROR/original represent the metrics for the Original dataset, while PRR/clean and ROR/clean denote the metrics for the clean dataset. As shown in Fig. \ref{fig12}, the values of PRR/clean and ROR/clean are significantly higher than those of PRR/original and ROR/original. This indicates that after removing biased data, even when using traditional ROR and PRR methods to calculate the strength of ADR, the correlation is higher compared to the original dataset, further highlighting the impact of biased data on the strength of ADR.

Specifically, for adverse reactions such as headache, tachycardia, sick, and pain, the clean dataset exhibited higher values in the ROR/clean metric. For cough, tachycardia, fever, and rash, the clean dataset show higher values in the PRR/clean metric. When analyzing the results, we observe that not all adverse reaction categories demonstrate significantly higher metric values in the clean dataset compared to the original. This is because certain adverse reactions may be more prevalent or severe among the patient population, resulting in higher reporting rates and priority levels in the original dataset already. Consequently, the data cleaning process may not significantly alter the metric values for these categories, as they have already approached or reached the statistical upper limit.

\section{Conclusion}
This paper proposes an efficient ADR signal prediction model called PFed-Signal, which centers around two innovative components: Pfed-Split and ADR-signal. Pfed-Split is a federated learning-based strategy for partitioning ADRs. ADR-signal, on the other hand, first identifies and removes biased data through federated learning and Euclidean distance to generate a clean dataset. It then introduces a Transformer-based detection model specifically designed to identify and reconstruct signal strengths on the clean dataset. Experimental results demonstrate that the PFed-Signal model significantly outperforms methods such as SVN, BCPNN, and RF in key metrics including accuracy, F1 score, and AUC. Future research will focus on further optimizing the Pfed-Split and ADR-signal components to enhance the model's generalization ability and improve its adaptability across different datasets.

%\nocite{*}
\bibliographystyle{IEEEtran}
\bibliography{main}
\end{document}